\documentclass{article}
\usepackage{spconf,amsmath,graphicx}
\usepackage{amsmath,amssymb}
\usepackage{cite}
\usepackage{color}
\usepackage{bbm}
\usepackage{subfigure}
\usepackage{multicol}
\usepackage{multirow}
\usepackage{array}
\usepackage{diagbox}
\usepackage{comment}
 
\newcommand{\etal}[1]{\textit{et al}\onedot}
\usepackage{url}

\def\etal{\emph{et al.}}

\title{ACTION ANTICIPATION WITH GOAL CONSISTENCY}
\name{Olga Zatsarynna and Juergen Gall}
\address{University of Bonn \& Lamarr Institute}
\begin{document}
%
\maketitle
\begin{abstract}
In this paper, we address the problem of short-term action anticipation, i.e., we want to predict an upcoming action one second before it happens. We propose to harness high-level intent information to anticipate actions that will take place in the future. To this end, we incorporate an additional goal prediction branch into our model and propose a consistency loss function that encourages the anticipated actions to conform to the high-level goal pursued in the video. In our experiments, we show the effectiveness of the proposed approach and demonstrate that our method achieves state-of-the-art results on two large-scale datasets: Assembly101 and COIN. 
\end{abstract}
\vspace{-0.1cm}
\section{Introduction}
\label{sec:intro}
Anticipation of human actions is a task that we naturally solve in various day-to-day situations. We anticipate movements of cars while crossing the road, predict the plot elements of a new movie and picture how someone will react to our own actions. For all of these scenarios, we are able to imagine the future and adjust our beliefs and behavior accordingly. Due to how ubiquitous the situations that require the ability for action anticipation are, it is crucial that the intelligent agents designed to operate among human beings get hold of this task. 

\begin{figure}[t!]
\begin{minipage}[b]{0.5\linewidth}
  \centering
  \includegraphics[scale=0.25]{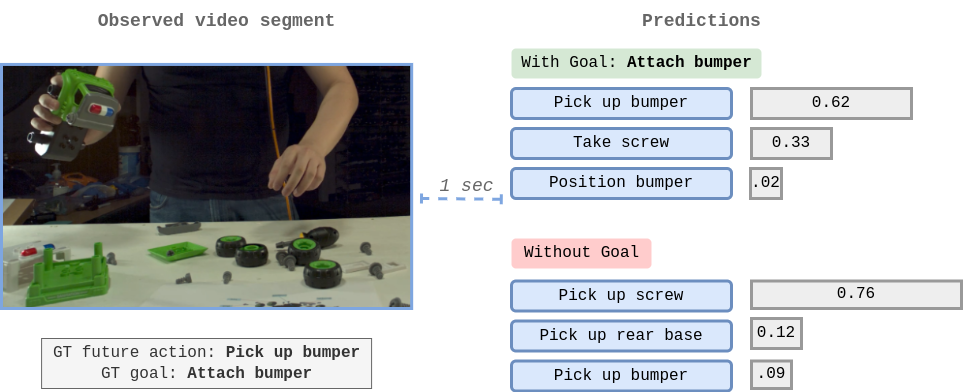}
\end{minipage}
\caption{\small To anticipate future actions, we propose to incorporate high-level intent prediction as part of our model. By relying on this information, the prediction model can filter out fine-grained actions that are not in correspondence with the pursued goal.}
\label{fig:intro}
\end{figure}

In this work we consider the setting of short-term anticipation, i.e., we want to anticipate a single action one second before it happens. Short-term anticipation has been addressed in different works~\cite{furnari2020rulstm, zatsarynna2021MMTCN, zatsarynna_2022_gcpr, nawhal2022anticipatr, memvit2022, girdhar2021anticipative, Gao_2017_BMVC, Vondrick_2016_CVPR, Liu_2020_ECCV, Dessalene_2021_forecasting, Miech_2019_CVPR_Workshops, Jain_2016_ICRA},  
that showed impressive performance on this task in question. Most of these approaches, however, directly predict future actions without taking into account what has driven humans to undertake these actions in the first place. Yet, we observe that understanding the intent behind actions can simplify the task of anticipation. This is because different high-level goals are associated with different subsets of lower-level actions required to complete them. For example, as shown in Figure~\ref{fig:intro}, if we know that the person's goal is to \textit{attach a bumper} to a toy vehicle, we can conclude that the upcoming action could be \textit{pick up bumper or position bumper} depending on the progress of the goal, but not \textit{pick up rear base, pick up cabin}. In this way, goal awareness reduces the number of valid options for future actions and thereby makes the task of anticipation simpler to solve. 
Motivated by this observation, we introduce in addition to the fine-grained action anticipation branch a separate goal prediction branch into our model. 
Nevertheless, simply forecasting both actions and their high-level goals independently does not explicitly ensure consistency between these two predictions. Therefore, we formulate an additional consistency loss that forces the fine-grained branch to predict actions that afford the completion of the pursued goal.
Overall, the contributions of this work can be summarized as follows:
\begin{itemize}
    \vspace{-0.25cm}
    \item We propose to harness high-level goal information to facilitate anticipation of future actions. We incorporate an additional goal prediction branch and propose a consistency loss to encourage alignment between predicted future actions and the underlying high-level intent.
    \vspace{-0.25cm}
    \item We demonstrate that our proposed approach achieves state-of-the-art results for the task of action anticipation on two large-scale procedural datasets: Assembly101 and COIN.
\end{itemize}

\vspace{-0.4cm}
\section{Related work}
\label{sec:related_work}

\textbf{Action Anticipation in Videos.} Works on action anticipation generally deal with one of the two established directions: long-term or short-term anticipation. Long-term anticipation works are focused on predicting multiple actions into the future with a forecasting horizon of several minutes. Short-term anticipation methods, on the other hand, focus on predicting only the next action a few seconds in advance. Both research directions have received increased attention in recent years due to the availability of new large-scale datasets~\cite{ Damen2021PAMI, sener2022assembly101, yin2018eye}.

Starting with long-term action anticipation, Abu Farha~\etal~\cite{Farha_2018_CVPR} introduced two anticipation approaches based on RNN and CNN networks. While the CNN network performed anticipation in one shot relying on a matrix representation of actions, the RNN model predicted actions and their length autoregressively and achieved superior results.
To avoid the accumulation of errors due to the autoregressive prediction, Ke~\etal~\cite{Ke_2019_CVPR} introduced a temporal convolutional time-conditioned network that anticipated all upcoming actions in one shot. Recently, Gong~\etal~\cite{gong2022future} and Nawhal~\etal~\cite{nawhal2022anticipatr} proposed two transformer-based~\cite{vaswani2017attention} architectures for long-term anticipation.

The second line of work focuses on anticipation of the next action several seconds before its onset. Vondrick~\etal~\cite{Vondrick_2016_CVPR} proposed to solve this task by regressing the representation of a single future frame and classifying it to get the future action prediction. Extending upon~\cite{Vondrick_2016_CVPR}, Gao~\etal~\cite{Gao_2017_BMVC} used an encoder-decoder network to process several observed frames and anticipate multiple future representations instead of just one. Another sequence-to-sequence approach was introduced by Furnari~\etal~\cite{furnari2020rulstm} - an RU-LSTM model consisting of two LSTM networks for past summarization and future action prediction respectively. Zatsarynna~\etal~\cite{zatsarynna2021MMTCN} proposed a temporal convolutional network to tackle the inefficiency of the approaches relying on the recurrent layers. In~\cite{sener2020temporal}, Sener~\etal\ presented a TempAgg model that used non-local-block~\cite{Wang2017NonlocalNN} attention to encode the observed video snippets at different temporal scales creating recent and spanning features used for both short-term and long-term anticipation. Several recent works~\cite{zhao2022testra, girdhar2021anticipative, memvit2022} proposed transformer-based architectures to allow for information flow between distant parts of the observed video segments, as well as enable spatial attention within individual video frames. In contrast to these approaches, our work focuses on harnessing information about action goals to predict future actions more accurately.

\textbf{Intention-driven Forecasting.} So far only very few works have addressed intention or goal-driven forecasting \cite{tanke2021intention,Mascaro_2023_WACV,debaditya2022latent}. These works define intention in a different way or address other tasks. Debaditya~\etal~\cite{debaditya2022latent} defines the goal `as the visual representation after performing the final action based on the procedure planning paradigm'. The approach thus aims to forecast visual features that are closer to the expected visual representation at the end of the sequence. Mascaro~\etal~\cite{Mascaro_2023_WACV} addresses long-term action anticipation and conditions a VAE on the high-level activity. Tanke~\etal~\cite{tanke2021intention} forecast the future actions ahead of time to generate smooth and plausible human motion sequences. 



\begin{figure}[h!]
\begin{minipage}[b]{0.5\textwidth}
  \centering
  \includegraphics[scale=0.23]{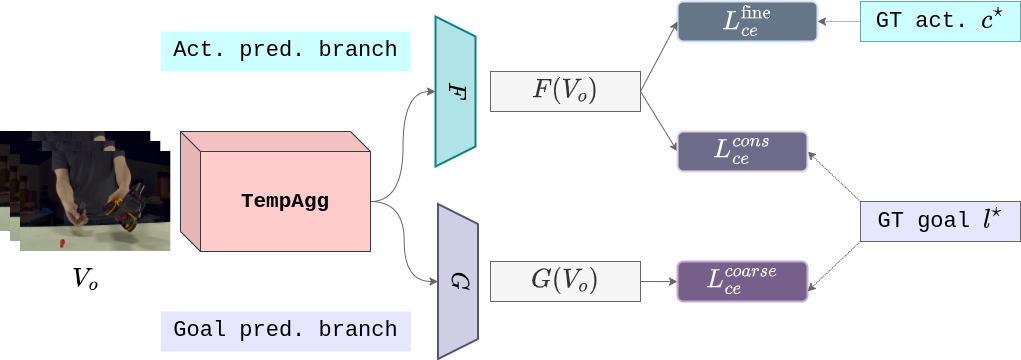}
\end{minipage}
\caption{\small Overview of our proposed approach. Our model contains two branches: an action branch that predicts future actions and a goal branch that predicts future goals. The goal prediction branch is trained using cross-entropy loss $L_{ce}^{coarse}$, while the action branch is trained with a combination of a cross-entropy $L_{ce}^{fine}$ and a consistency loss $L_{ce}^{cons}$.}
\label{fig:model}
\vspace{-0.3cm}
\end{figure}

\vspace{-0.2cm}
\section{Method}
\vspace{-0.1cm}
\label{sec:method}
We start by formally defining the task of short-term action anticipation in Section~\ref{subsec:task}. We then describe our proposed approach in Section~\ref{subsec:model}. 
\vspace{-0.3cm}
\subsection{Task}
\vspace{-0.1cm}
\label{subsec:task}
Following~\cite{sener2020temporal, furnari2020rulstm}, we define the task of action anticipation as follows: given the observed video segment $V_o$ that precedes the action of interest, we want to predict its label $T_a$ seconds before the onset. In our work, we consider the anticipation time of one second, i.e. $T_a = 1$.
\vspace{-0.2cm}
\subsection{Model}
\vspace{-0.1cm}
\label{subsec:model}
\textbf{Network branches.} To address the above-defined task, we propose to harness information about the goals behind the upcoming actions. Intuitively, knowledge of the pursued goal simplifies the task of anticipation by constraining the number of valid future action choices.
To make use of the goal information, as shown in Figure~\ref{fig:model}, the network consists of two prediction branches: a fine-grained action anticipation branch $F$ and a goal prediction branch $G$. 
While the fine-grained action anticipation branch is trained for the final task of future action anticipation, the goal branch learns to predict the goals behind these actions. These two branches share the same backbone, namely TempAgg~\cite{sener2020temporal}.

Formally, the fine-grained action branch $F$ receives the observed video segment $V_o$ as input and outputs the probability distribution of the next action $F(V_o) \in \mathbb{R}^{|C|}$, where $C$ is the set of all fine-grained actions. Similarly, the goal branch $G$ takes $V_o$ as input and outputs probability distribution of the goal $G(V_o) \in \mathbb{R}^{|L|}$, where $L$ is the set of all possible goal classes. Both branches are optimized using the cross-entropy loss:
\begin{align}
    L_{ce}^{fine} &= -\sum_{n}\sum_{c \in C} \mathbbm{1}(c = c^*_n) \log(F(V_o^n)_c), \\
    L_{ce}^{goal} &= -\sum_{n}\sum_{l \in L} \mathbbm{1}(l = l^*_n) \log(G(V_o^n)_l),
\end{align}
where $n$ is the batch index, $F(V_o^n)_c$ and $G(V_o^n)_l$ are $c^{th}$ and $l^{th}$ element of the corresponding probability distribution, and $c^*_n$ and $l^*_n$ are ground-truth action and goal labels, respectively.

\textbf{Consistency loss.} Simply incorporating a separate goal prediction branch into the model does not explicitly ensure that its predictions and the predictions of the action branch are aligned with each other. Here, by alignment, we mean that a fine-grained action will lead to a progress or completion of a given goal. For example, a fine-grained action \textit{screw chassis} is aligned with the goal \textit{attach chassis}, while an action \textit{screw water tank} is not. First, to understand which actions align with which goals, we compute how often individual goals and actions occur together in the training set. Then, to enforce the alignment on the model's fine-grained action predictions, we make use of the previously obtained co-occurrence statistics to formulate the consistency loss. 
More specifically, we first map fine-grained action predictions to probability distributions over the goal classes. To this end, we use the joint probability marginalization formula, where we substitute the action probabilities by the predictions of the fine-grained action branch:
\begin{align}
    \hat{G}(V_o^n)_{l} &= \sum_{c \in C}P(l|c)F(V_o^n)_{c} .
    \label{remap}
\end{align}

To estimate the conditional distribution $P(l|c)$, we collect the action-goal co-occurrence matrix $M \in \mathbb{R}^{|L|\times|C|}$, where entry $M(l,c)$ stores the number of times goal $l$ and fine-grained action $c$ occurred together in the training set:
\begin{align}
    M(l,c) = |\{n \in \{1, \dots, N\} | c^*_n = c \land l^*_n = l\}|, \\
    \nonumber\forall c \in C, \forall l \in L .
\end{align}

Here, $N$ is the total number of training examples in the dataset. Having acquired the co-occurrence matrix, we first approximate the joint probability distribution $P(l,c)$ and then obtain $P(l|c)$:
\begin{align}
    P(l,c) &\approx \frac{M(c,l)}{\sum_{c', l'}M(c',l')} , \\
    P(l|c) &= \frac{P(l,c)}{\sum_{l'}P(l',c)} .
\end{align}

Finally, we compute our consistency loss as the cross-entropy loss between the obtained remapped goal distributions \eqref{remap} and the true goal labels.
Formally:
\begin{align}
    L_{ce}^{cons} = -\sum_{n}\sum_{l \in L} \mathbbm{1}(l = l^*_n) \log(\hat{G}(V_o^n)_l) .  
\end{align}
This loss ensures that the predicted actions $F(V_o^n)$ are aligned with the underlying goals $l^*_n$ according to the predefined action-goal hierarchy given by the conditional probability.

\textbf{Final loss.} To conclude, we train our network with the linear combination of the previously discussed branch-wise and consistency losses:
\begin{align}
L = L_{ce}^{fine} + L_{ce}^{goal} + \lambda_{cons}L_{ce}^{cons},
\end{align}
where $\lambda_{cons}$ weights the consistency loss. 

\vspace{-0.25cm}
\section{Experiments}
\label{sec:experiments}

\begin{table*}[t!]
\centering
\scriptsize
\begin{tabular}{|c|c|c|c||c|c|c|c|c|c|c|c|c|c|}
    \hline
    \multirow{3}{*}{Model}& \multicolumn{12}{c|}{M. Top-5 Rec.\%} & \multirow{3}{*}{Params} \\
    \cline{2-13}
    & P.V. ACT. & P.V. NOUN & P.V. VERB & \multicolumn{3}{c|}{M.V. ACT.} & \multicolumn{3}{c|}{M.V. NOUN} & \multicolumn{3}{c|}{M.V. VERB} & \\
    \cline{2-13}
    & Overall & Overall & Overall & Overall & Unseen & Tail & Overall & Unseen & Tail & Overall & Unseen & Tail &  \\
    \hline
    \hline
    TempAgg~\cite{sener2022assembly101} & 8.19 & 25.59 & 54.61 & 8.53 & 8.34 & 3.94 & 26.27 & 23.00 & 25.93 & 59.11 & 58.77 & 53.10 & 207M \\
    \hline
    \hline
    No goal & 8.74 & 26.89 & \textbf{55.85} & 9.53 & 8.77 & 5.00 & 26.94 & \underline{23.40} & 26.14 & \underline{59.87} & \textbf{59.73} & \underline{53.41} & 207M \\
    \hline
    Ours (1 goal) & \underline{10.39} & \underline{27.50} & 54.59 & \underline{11.29} & \underline{9.69} & \underline{6.71} & \underline{27.66} & 23.32 & \underline{26.84} & 58.40 & 58.17 & 52.59 & +330.0K \\ 
    \hline
    Ours (2 goals) & \textbf{10.64} & \textbf{27.63} & \underline{55.82} & \textbf{12.07} & \textbf{10.81} & \textbf{7.68} & \textbf{28.38 }& \textbf{23.64} & \textbf{27.78} & \textbf{60.04} & \underline{59.63} & \textbf{53.87} & +61.47K \\
    \hline
\end{tabular}
\caption{\small Action anticipation results on the Assembly101 validation set. \textit{P.V.} and \textit{M.V.} stand for per-view and multi-view evaluation, respectively. In the first case, different views of the same video sequence are considered as separate examples, while in the second case, only one prediction per video sequence is made by averaging results over all the views associated with it.}
\label{tab:res_assembly}
\end{table*}

\vspace{-0.25cm}
\subsection{Datasets and Evaluation}
\vspace{-0.1cm}
Since our approach relies on a predefined action-goal hierarchy,  we use two procedural activity datasets that contain hierarchical action annotations: Assembly101~\cite{sener2022assembly101} and COIN~\cite{tang2019coin}. 

\textbf{Assembly101} is a large-scale dataset that contains 362 recordings of 15 toy-vehicle assembly and disassembly sequences shot from 12 different viewpoints. The videos are annotated with 1M fine-grained and 100K coarse action segments, that we use as fine-grained actions and goals, respectively. Fine-grained segments span 1380 action classes composed of 90 objects and 24 verbs, while coarse actions span 202 action classes formed by 11 verbs and 69 objects. Assembly101 is divided into training, validation, and test splits. At the time of writing, the test set was not available, thus in our work we report results on the validation split. Following~\cite{sener2022assembly101}, we additionally provide results on two subsets of validation examples - \textit{Tail} and \textit{Unseen} - that contain video segments with tail action classes and toys unseen during training time, respectively.

\textbf{COIN} consists of 11827 videos that were collected from Youtube. The videos depict 180 high-level tasks and are annotated with 46354 action segments from 778 lower-level action classes. In our experiments, we regard video-level task annotations as goal actions, while segment-level action annotations as fine-grained actions. Training and testing splits contain subsets of 9030 and 2797 videos, respectively.

For performance evaluation, we used Class-Mean Top-5 Recall following~\cite{sener2022assembly101} to account for the uncertainty of future predictions. For Assembly101, we report action, noun, and verb recall, while for COIN only action recall.

\vspace{-0.25cm}
\subsection{Implementation Details}
\label{ssec:subhead}
\vspace{-0.1cm}
For our experiments, we adopted the TempAgg model from \cite{sener2022assembly101} as the baseline and made use of RGB features provided by~\cite{sener2022assembly101, tang2019coin} for the corresponding datasets. In addition to the already existing branches, we incorporated a separate goal prediction branch that operates on the spanning features, similar to~\cite{sener2020temporal}. For training, we used batch size 64 instead of 32. It improves the results as shown in the first two rows of Table~\ref{tab:res_assembly}\footnote{We follow the official protocol \url{https://github.com/assembly-101/assembly101-action-anticipation/tree/main/tempAgg-action-anticipation}, which differs from \cite{sener2022assembly101}.}. 

\vspace{-0.25cm}
\subsection{Results}
\label{subsec:results}
\vspace{-0.1cm}
We present the results of our method on Assembly101 and COIN  in Table~\ref{tab:res_assembly} and~\ref{tab:res_coin}, respectively. On both datasets, our method achieves improvements over the previously proposed TempAgg~\cite{sener2022assembly101} approach. We note that the main focus is on the performance of action anticipation, while verb and noun predictions are secondary. On Assembly101 our model outperforms TempAgg~\cite{sener2022assembly101} and our baseline (No goal) on the overall set of actions by 1.65\% and 1.76\% in the per-view and multi-view settings accordingly, while on COIN our approach achieves 0.54\% improvement. On Assembly101, we further experimented with extending the model with one more branch (Ours (2 goals)), that would predict an even higher-level video sequence goal: assembly/disassembly of a particular toy type \textit{(i.e. assembly truck, disassembly SUV)}. The consistency loss for this goal type is computed analogously to $L_{ce}^{cons}$. This extension brought further $0.25\%$ and $0.78\%$ improvements in the overall action recall for the per-view and multi-view settings, respectively. The increase of the number of model parameters by the goal prediction branches is very small as shown in Table~\ref{tab:res_assembly} and~\ref{tab:res_coin}. 
Since the goal branches are discarded at the test time, the inference cost remains the same.

\begin{table}[h!]
    \centering
    \scriptsize
    \begin{tabular}{|c|c|c|c|}
        \hline
         Method & ACT. M. Top-5 Rec & Params \\
         \hline
         No goal & 13.39 & 61.072M \\
         Ours & \textbf{13.93} & + 369.0K \\
        \hline
    \end{tabular}
    \caption{\small Action anticipation results on COIN validation set.}
    \label{tab:res_coin}
\end{table}
\vspace{-0.6cm}

\subsection{Ablation}
\vspace{-0.1cm}
In this section, we present the results of the ablation studies for our method. We inspect the impact of each loss term, ablate the formulation of the consistency loss, and analyze the effect of the loss weight $\lambda_{cons}$.

\begin{table}[h!]
    \scriptsize
    \centering
    \begin{tabular}{|l|c|c|}
        \hline
         & \multicolumn{2}{c|}{ACT. M. Top-5 Rec.} \\
         \hline
         Method & COIN & Assembly101 (P.V.)\\
         \hline
         $L_{ce}^{fine}$ & 13.39 & 8.74 \\
         $L_{ce}^{fine} + L_{ce}^{goal}$ & 13.57 & 9.09 \\
         $L_{ce}^{fine} + L_{ce}^{goal} + L_{ce}^{cons}$ & \textbf{13.93} & \textbf{10.39} \\
         \hline
    \end{tabular}
    \caption{\small Ablation of the final loss components.}
    \label{tab:abl_components}
\end{table}

\textbf{Loss components.} The loss function for our model consists of three terms: $L_{ce}^{fine}$, $L_{ce}^{goal}$ and $L_{ce}^{cons}$. To analyze how individual terms impact the final performance, we train separate models with different combinations of these loss terms. The results of this experiment are presented in Table~\ref{tab:abl_components}. Using only the fine-grained action anticipation loss $L_{ce}^{fine}$ is always required and is equivalent to just applying the TempAgg~\cite{sener2022assembly101} model. Adding the goal prediction loss improves the performance by $0.18\%$ and $0.35\%$ for COIN and Assembly101, respectively. Further extending the final loss function with the consistency loss $L_{ce}^{cons}$ results in additional improvement of $0.36\%$ and $1.30\%$ accordingly. This shows the benefit of incorporating the additional goal prediction and consistency loss into the training loss function.

\textbf{Consistency loss formulation.} For computing the consistency loss, we make use of the ground-truth goal labels during training. Another possibility is to harness the predictions made by the goal branch instead. For comparison, we replace the cross-entropy loss by the KL divergence between the predictions of the goal branch and the remapped fine-grained action predictions. 
We show the results of this experiment in Table~\ref{tab:abl_formulation}.  We observe that the KL loss improves the performance compared to not using a consistency loss, but it is inferior to training with ground-truth goal labels. This result is intuitive since predicted goal distributions can be noisy as opposed to the ground-truth labels.
\begin{table}[h!]
    \centering
    \scriptsize
    \begin{tabular}{|l|c|c|}
        \hline
        & \multicolumn{2}{c|}{ACT. M. Top-5 Rec.} \\
        \hline
        Consistency Loss & COIN & Assembly (P.V.)  \\
        \hline
        Predicted & 13.65 & 9.89 \\
        Ground-truth (Ours) & \textbf{13.93} &\textbf{10.39} \\
        \hline
    \end{tabular}
    \caption{\small Ablation of the consistency loss formulation.}
    \label{tab:abl_formulation}
\end{table}

\textbf{Consistency loss weight.} The consistency term in the final loss function is weighted by a factor $\lambda_{cons}$. We evaluate the impact of this hyper-parameter in Table~\ref{tab:abl_lambda}. We observe that lower values of $\lambda_{cons}$ perform better for the COIN dataset, while higher values perform better for Assembly101. The reason is that Assembly101 has a higher ratio of fine-grained actions to goals than COIN, which is compensated by a higher $\lambda_{cons}$.


\begin{table}[h!]
    \centering
    \scriptsize
    \begin{tabular}{|c|c|c|c|c|}
        \hline
        {Dataset}& \multicolumn{4}{c|}{$\lambda_{cons}$ / ACT. M. Top-5 Rec.} \\
        \hline
        \multirow{2}{*}{COIN} & 0.1 & 0.5 & 1.0 & 2.5 \\
        \cline{2-5}
        & 13.63 & \textbf{13.93} & 13.91 & 12.85  \\
        \hline
        \hline
        \multirow{2}{*}{Assembly101} & 1.0 & 2.5 & 5.0 & 10.0 \\
        \cline{2-5}
        & 9.65 & 10.12 & \textbf{10.39} & 10.24 \\
        \hline
    \end{tabular}
    \caption{\small Ablation of the consistency loss weight.}
    \label{tab:abl_lambda}
    \vspace{-0.5cm}
\end{table}

\section{Conclusion}
\label{sec:conclusion}
\vspace{-0.1cm}
In our work, we proposed to harness intent information to perform action anticipation by extending the model with a goal prediction branch and computing a goal-action consistency loss. We demonstrated that our proposed approach achieves stat-of-the-art results on two large-scale procedural activity datasets on the task of action anticipation.

\newpage
\bibliographystyle{IEEEbib}
\bibliography{refs}

\end{document}